\documentclass[conference]{IEEEtran}
\IEEEoverridecommandlockouts
\usepackage{cite}
\usepackage{amsmath,amssymb,amsfonts}
\usepackage{algorithmic}
\usepackage{graphicx}
\usepackage{textcomp}
\usepackage{xcolor}
\usepackage{longtable, array, booktabs}
\def\BibTeX{{\rm B\kern-.05em{\sc i\kern-.025em b}\kern-.08em
    T\kern-.1667em\lower.7ex\hbox{E}\kern-.125emX}}
\IEEEoverridecommandlockouts\IEEEpubid{\makebox[\columnwidth]{978-1-6654-
9653-7/22/\$31.00~\copyright~2022 IEEE \hfill}
\hspace{\columnsep}\makebox[\columnwidth]{ }}

\begin{document}

\title{Multimodal Feature Extraction for Memes Sentiment Classification}

\author{\IEEEauthorblockN{Sofiane Ouaari\IEEEauthorrefmark{1}, Tsegaye Misikir Tashu\IEEEauthorrefmark{1} \IEEEauthorrefmark{3} and
Tom\'a\v{s} Horv\'ath\IEEEauthorrefmark{1}\IEEEauthorrefmark{2}}
\IEEEauthorblockA{\IEEEauthorrefmark{1} Department of Data Science and Engineering, Faculty of Informatics,
                         ELTE - E\"otv\"os Lor\'and University\\
                        P\'azm\'any P\'eter s\'et\'any 1/C, 1117 Budapest, Hungary }
\IEEEauthorblockA{\IEEEauthorrefmark{2}Institute of Computer Science,  Faculty of Science, Pavol Jozef \v{S}af\'arik University,\\ Jesenn\'a 5, 040 01 Ko\v{s}ice, Slovakia}
\IEEEauthorblockA{\IEEEauthorrefmark{3}College of Informatics, Kombolcha Institute of Technology, Wollo University, \\ 208 Kombolcha, Ethiopia}

Email: \IEEEauthorrefmark{1}dfmyzk@inf.elte.hu,
\IEEEauthorrefmark{1}misikir@inf.elte.hu,
\IEEEauthorrefmark{1}tomas.horvath@inf.elte.hu
}

\maketitle

\begin{abstract}

In this study, we propose feature extraction for multimodal meme classification using Deep Learning approaches. A meme is usually a photo or video with text shared by the young generation on social media platforms that expresses a culturally relevant idea. Since they are an efficient way to express emotions and feelings, a good classifier that can classify the sentiment behind the meme is important. To make the learning process more efficient, reduce the likelihood of overfitting, and improve the generalizability of the model, one needs a good approach for joint feature extraction from all modalities. In this work, we proposed to use different multimodal neural network approaches for multimodal feature extraction and use the extracted features to train a classifier to identify the sentiment in a meme.

\end{abstract}


\section{Introduction}

A meme is a type of satire that uses references from popular culture to reveal a hidden message. It is an effective technique for collecting thoughts, emotions, and behaviors in an easily transferable way, making them ideal for rapid dissemination. According to the Q1 2018 YPulse Trend Report survey \cite{ypulse}, which includes news/insights and research on Generation Z and Millennials, more than half of young consumers send memes weekly, and 30\% send them daily. The work of Iloh \cite{iloh2021culture} has shown how memes can be effectively incorporated into qualitative research and publications for social scientists and anthropologists.

However, most of the tools available on social media can be used by some people to spread and send negative vibes and hateful statements, and memes are no exception. The work by \cite{phdthesis_holland} examines the social impact, both positive and negative, of memes as a new form of communication and concludes that Internet memes are often very influential and affect people in various ways that may not be immediately apparent.
 
Most studies dealing with meme classification have used a single feature selection and extraction approach for each modality used to train the classifier {\cite{keswani2020iitk,gupta2020dsc}}. Since the inputs to the models are multimodal and one modality complements the other modality, an approach that jointly learns and extracts a multimodal feature is important. Since there are two input modalities in a meme, namely the image and the text, this work proposed and explored how different multimodal deep learning models can help extract important features and patterns to detect the overall sentiment of a meme on a three-level scale [negative-neutral-positive].
According to the experimental results, the multimodal feature extraction approach was a good choice for training a classifier and gave efficient results on the testing set. The rest of the paper is structured as follows. Section II presents an overview of multimodal learning and related works. Section III describes the dataset used in the study. The multimodal feature extraction approaches, ML models, and experimental settings are presented in Section IV. \textbf{Section V} presents the experimental results and discussions of the performed experiments. Section VI concludes the study.

\section{Literature Review}
\subsection{What is multimodal learning}

In the last decade, deep learning approaches have seen exponential development due to the overwhelming amount of data and growing technical offerings in the hardware market. However, most of the state-of-the-art models are mainly unimodal oriented and use data from a single modality as input. The ultimate goal of artificial intelligence research is to mimic the human brain, which can process multiple input modalities simultaneously, as much as possible.

Our daily lives consist of multiple modalities of different types, from the things we see to the things we taste to the sounds we hear, and we process them continuously. Multimodal machine learning explores different model architectures that can better represent and understand multiple inputs, even though they differ in encoding and are heterogeneous in nature.

\subsection{Multimodal Learning Taxonomy}

Baltrusaitis et al. {\cite{baltruvsaitis2018multimodal}} pointed out five challenges and concepts that exist in a multimodal learning context. These are:

\begin{enumerate} 

\item \textbf{Representation:}  One of the most difficult aspects of multimodal data is summarizing the information from the given modalities and determining the best representation approache that will help the model perform better on the given task. In multimodal challenges, the ability to describe data in a meaningful way is critical and forms the basis of any model. 

\item \textbf{Translation:}  It is about generating an instance of the input but in a different modality. The best-known tasks in this area are spoken language translation, image-driven translation, and video-driven translation, which exploit the audio and visual modalities respectively {\cite{sulubacak2020multimodal}}. 

\item \textbf{Alignment:} Finding connections and correspondences between subcomponents of examples from two or more modalities are called multimodal alignment.

\item \textbf{Fusion:} To improve the generalization performance of complex cognitive systems, it is essential to capture and fuse an appropriate set of useful features from the available modalities. 

\item \textbf{Co-learning:} In real-world tasks, one or more modalities are often found to be absent; noisy, insufficiently annotated, with unreliable labels, in short supply in training or testing, or both. Multimodal co-learning is a learning paradigm that addresses this problem. Knowledge transfer between modalities, including their representations and predictive models, helps model one modality (resource-poor) by leveraging the knowledge of another modality (resource-rich).
\end{enumerate}

\subsection{Multimodal Learning for Memes Sentiment Analysis} 

The problem we are working on in this paper falls under \emph{task A} of SemEval-2020 Task 8 " Memotion Analysis" {\cite{chhavi2020memotion}}, where detailed information about the dataset can be found in \textbf{Section III}.

A comparative study between different unimodal and multimodal architectures in the sentiment classification for memes was made by {\cite{keswani2020iitk}} and {\cite{gupta2020dsc}}. The work of {\cite{keswani2020iitk}} obtained the best results with a "text-only feed forward neural network (FFNN)" model, using Word2Vec {\cite{mikolov2013efficient}} for representing each word as a vector ($1\times300$) and then averaging the Word2Vec {\cite{mikolov2013efficient}} embeddings of words used for each caption. This unimodal approach allowed {\cite{keswani2020iitk}} to score first in the related SemEval2020 task 8 "Memotion Analysis" challenge with a macro F1 score of \textbf{35.47\%}.  \\
The comparative study of {\cite{gupta2020dsc}} achieved its best result with a bimodal model by combining ResNet {\cite{tai2017image}} for the visual input and RoBERTa (Robust Optimized BERT pre-training Approach) {\cite{liu2019roberta}}, which is an approach that better trains and optimizes the original BERT {\cite{devlin2018bert}}, on text.

\subsection{Multimodal Feature Extraction}

A good common practice in the machine learning sphere is to apply feature extraction {\cite{khalid2014survey}}, which is a step in the dimensionality reduction process, by reducing the large set of raw data and keeping the most important features which efficiently describes the dataset. For unimodal cases, PCA and Kernel PCA {\cite{cao2003comparison}} were used to extract features for the face recognition task {\cite{ebied2012feature}} and further compared linear versus non-linear methods for feature extraction. In facial emotion detection, {\cite{baltruvsaitis2016openface}} proposed the open-source software "OpenFace" being able to represent facial landmarks, head pose estimations, facial action unis and eye-gaze estimations. Data feature extraction for sequential data was also developed such as "OpenSmile" {\cite{eyben2015opensmile}} which generates low-level audio features relevant in the emotional speech detection framework. 

In addition, some work has also been done to perform feature extraction in multitask problems and in multimodal tasks. Guillaume et al. {\cite{obozinski2006multi}} proposed a type of joint regularization method for feature selection across tasks by using both L1 and L2 norms and forcing the different predictors of different tasks to have close representations of their learned patterns. Yong et al. {\cite{luo2015large}} extended the complexity by developing a novel large margin multimodal multitask feature extraction (LM3FE) framework for handling multimodal features for image classification by learning the feature extraction matrix for each modality and the combination coefficients to handle correlated features and leverage the complementarity between them to reduce redundancy.

\section{Dataset} 

The dataset we used in this study was provided by Chhavi et al {\cite{chhavi2020memotion}} at the SemEval-2020 challenge called \emph{Memotion Analysis}.  SemEval is a series of international research workshops on natural language processing (NLP). The goal is to improve state-of-the-art semantic analysis and support the creation of high-quality annotated datasets for a variety of increasingly difficult semantic challenges in natural language. Each year, the workshop includes a series of collaborative challenges where computational semantic analysis systems from different teams are presented and compared. The dataset contains 6992 instances for the following three tasks: Sentiment Classification (Task A), Humor Classification (Task B), and Scales of Semantic Classes (Task C). The authors collected the data by selecting 52 globally known categories and using a browser extension tool to save memes on these topics from the Google image search engine. They then filtered the memes by keeping only those in English that had a clear background image with embedded text. They also used the Google Vision OCR APIs to extract the text from these memes as a separate input modality. 

As mentioned in \textbf{section II }, in this work we will address \emph{task A} to improve the performance of the challenge where the baseline macro F1 score was set to \textbf{21.76\%}. The models used and the experimental methodology employed are presented in \textbf{section IV }. The distribution of sentiment classes in the data set is presented in {table \ref{tab1}}.

\begin{table}[htbp]
\caption{Labels Distribution}
\begin{center}
\begin{tabular}{|c|c|c|}
\hline
\textbf{Sentiment} & \textbf{Number of Instances} & \textbf{Percentage}\\ 
\hline
Positive & 4160 & 59.5\% \\
\hline 
Neutral & 2201 & 31.5\% \\ 
\hline 
Negative & 631 & 9.0\% \\
\hline
\end{tabular}
\label{tab1}
\end{center}
\end{table}

\section{Methodology}

In this section, we will explain and present the different steps used to preprocess the data, the implementation of the multimodal feature extraction approaches, and the classical supervised machine learning classifiers used for sentiment classification. 

\subsection{Preprocessing}
\begin{itemize}
    \item \textbf{Text preprocessing:}  During text preprocessing, the following tasks were performed: Tokenization, removal of punctuations, determiners, and prepositions; transformation to lowercase, and removal of stop words. Then, we stemmed the words to their root. Finally, we padded all the text entries to make them have the same length.
    
    \item \textbf{Image preprocessing:} The memes images presented in the dataset have different shapes. \emph{OpenCV} was used to uniformly reshape the images into  224$\times$224$\times$3 dimensions. We then  \emph{MinMaxScaling} was also used for input normalization. 
\end{itemize}

\subsection{Multimodal Feature Extraction} 

We now address the various multimodal architectures proposed and used in the experiment. These models are trained with the goal of using the joint representation space as learned features that are fed into classical ML models to perform sentiment classification. 
\begin{itemize} 

    \item \textbf{Multi-Embedding Model}: an embedding layer is used in natural language processing to assign a vector representation with a fixed dimension to each word. In this architecture, we will consider two types of word embeddings. 
    \begin{itemize} 
        \item \textbf{Glove} stands for "Global Vectors for Word Representation" {\cite{pennington2014glove}}. Glove learns by constructing a co-occurrence matrix that counts how often a word occurs in a context. Since it is a gigantic matrix, a global matrix factorization called \emph{Latent Semantic Analysis} is applied.
        \item \textbf{Fasttext{\cite{joulin2016bag}} }, unlike Glove which considers the word level as the smallest unit, Fasttext uses n-gram characters as the smallest unit. For example, if we take the word "apple" with n=3, the Fasttext representation of this word is $ <$ap, app, ppl, ple, le$>$. This helps to capture the meaning of shorter words and allows embeddings to understand suffixes and prefixes. The biggest advantage of using Fasttext is that it generates better word embeddings for rare words or even words not seen during training since the n-gram character vectors are shared with other words.
\end{itemize}
    
    \begin{figure}
    \centering
    \includegraphics[width=\linewidth]{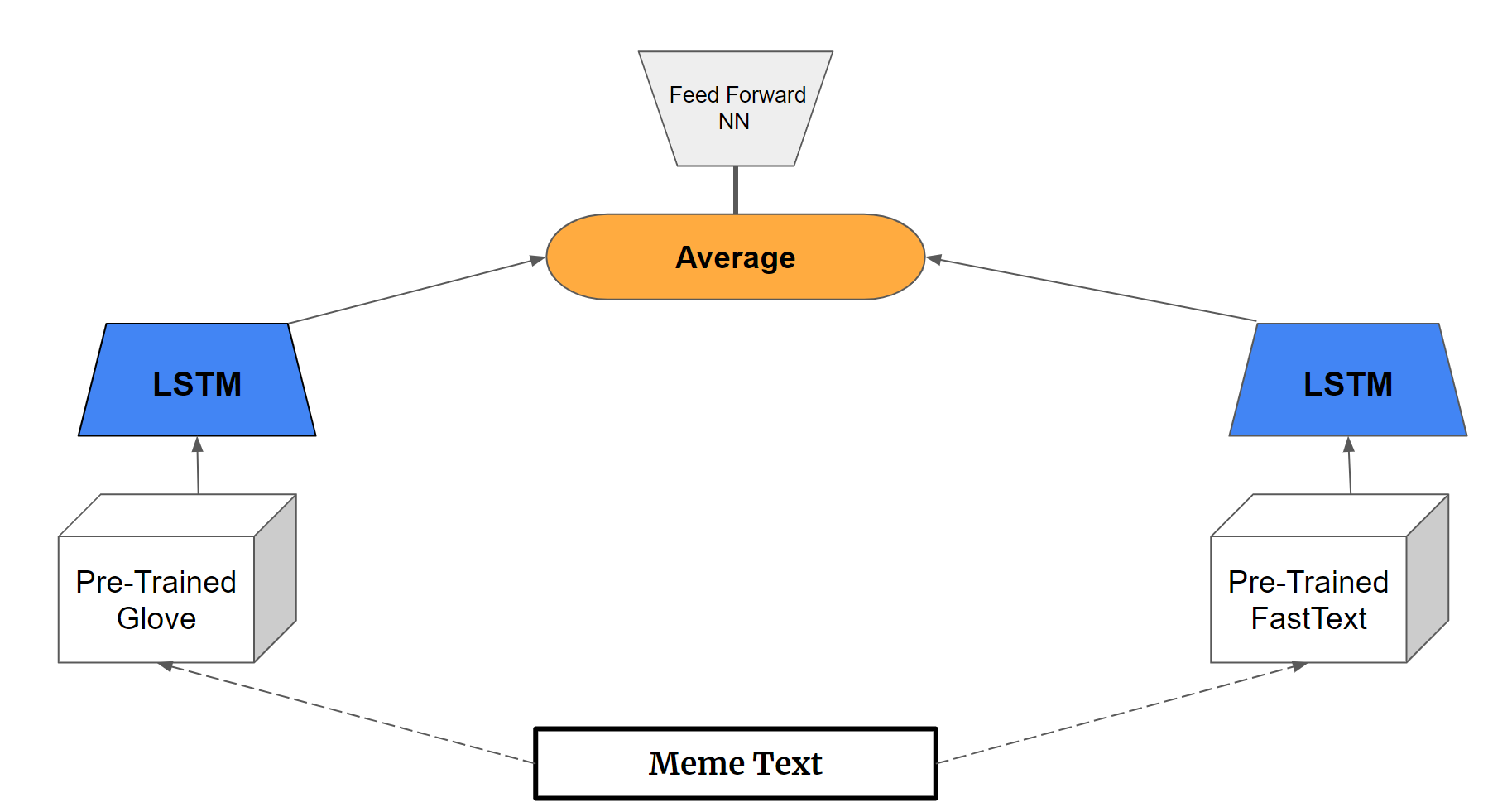}
    \caption{Multi-Embedding Architecture}
    \label{fig:multi_embedding_architecture}
\end{figure}

Multi-embedding is structured as shown in {fig \ref{fig:multi_embedding_architecture}}. Each text is represented by a pre-trained embedding layer of Glove {\cite{pennington2014glove}} and Fasttext {\cite{joulin2016bag}} and is routed in parallel to a Long Short Term Memory (LSTM) {\cite{hochreiter1997long}} block of size 256. We then averaged the outputs of the two LSTMs and feed them into a feed-forward neural network.

\item \textbf{Bimodal Autoencoder}: In \textbf{Section II.B}, we defined how input \emph{representation} is an important  and challenging  property of multimodal learning. A bimodal autoencoder is an architecture where shared representation is highly important. Autoencoders {\cite{bank2020autoencoders}} are unsupervised deep learning-based algorithms that reduce the number of dimensions in the data in order to encode it. Once the data was encoded using the autoencoders, it is decoded again on the other side. The system is complete when the data matches on both sides of the encoding.\\ 
 
Following this definition, we can extend it to a bimodal autoencoder, where we consider two input modalities, and the encoding is merged in a joint latent space where both are decoded and recreated. Hadeer et al.{\cite{sayed2021bimodal}} proposed a "Bimodal Variational Autoencoder (BiVAE)" model for the fusion of audiovisual features and their obtained experimental results show the superiority of the proposed model (BiVAE) for audiovisual feature fusion over the state-of-the-art models. \\

\begin{figure}
    \centering
    \includegraphics[width=\linewidth]{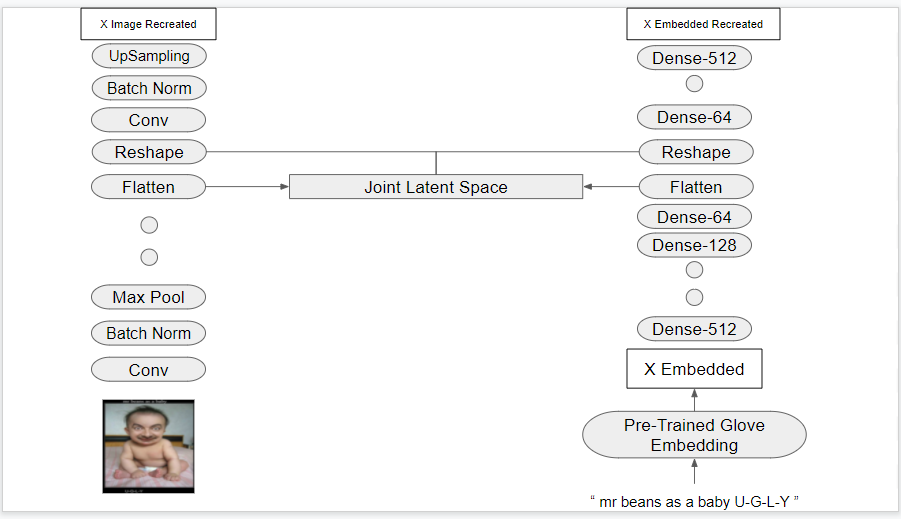}
    \caption{Bimodal Text-Image Autoencoder}
    \label{fig:bimodal_text_image_autoencoder}
\end{figure}

Fig  \ref{fig:bimodal_text_image_autoencoder} shows the implemented bimodal autoencoder with our meme data. For our supervised sentiment classification task, we will use the joint latent space as the set of learned features and adapt it to a different classification model.
\item{\textbf{Residual Multimodal}}: ResNet {\cite{he2016deep}} is originally used for image classification and allows building a deeper network due to the presence of \emph{skip connection} approach. 
\\ 
The idea of residual blocks can also be applied to multiple inputs. \cite{kim2016multimodal,jimaging7080157,Tashu} proposed Multimodal Residual Networks (MRN) for multimodal residual learning of visual question answering, which extends the idea of deep residual learning.
    
    \begin{figure}
    \centering
    \includegraphics[width=\linewidth]{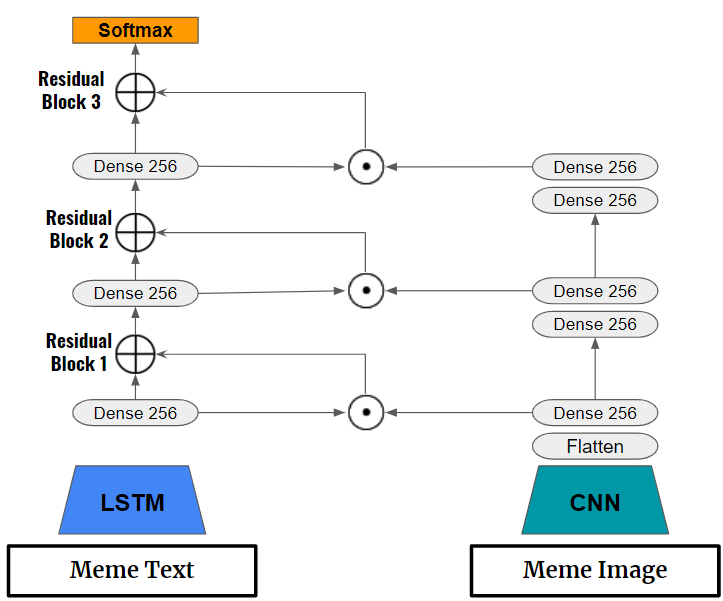}
    \caption{Multi Residual Network}
    \label{fig:multi_res_network}
\end{figure}

\end{itemize}

As shown in {fig \ref{fig:multi_res_network}}, we first pass the tokenized text data to an LSTM layer and the image to a convolutional layer and then flatten it. After passing them through a dense layer in parallel, we apply an element-wise multiplication \textbf{$\odot$} to it and create a residual block by performing an addition operation \textbf{$\oplus$} and loop over the same set of operations. In the feature extraction process, we will use the output of the residual blocks.

\subsection{Experimental setting}

We split the data into train/validation/test following with the ratio of 75/10/15\% respectively. We trained both \textbf{Multi-Embedding} and \textbf{Residual Multimodal} for the classification task, while the \textbf{Bimodal Autoencoder} was trained for performing a data reconstruction task. We monitored the \emph{validation loss} during training by using  \emph{Early Stopping}  to check if no more improvement is being noticed for a duration of five (5) epochs.

\subsection{ML Models}

\begin{figure}
    \centering
    \includegraphics[width=\linewidth]{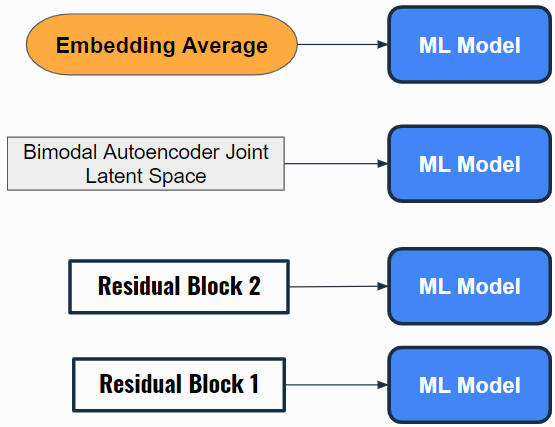}
    \caption{Multimodal Feature Extraction Pipeline}
    \label{fig:multi_fea_pipeline}
\end{figure}

After training the multimodal models mentioned in the previous section, we used their hidden representations as extracted features and train classical ML approaches for predicting the correct sentiment classes as shown in {fig \ref{fig:multi_fea_pipeline}}.  \\ 
Due to the multi-class nature of the memes' sentiment classification task, we adopted a decomposition-based logic of One-Vs-All {\cite{rifkin2004defense}}. We compared the performance of the following models: LinearSVC {\cite{rosipal2003kernel}}, K-Nearest Neighbors {\cite{laaksonen1996classification}}, Decision Trees{\cite{rokach2005decision}}, Random Forest {\cite{cutler2012random}} and Gradient Boosting Classifier {\cite{mayr2014evolution}}.

\section{Experimental Results}

As discussed in \textbf{Section IV.B},  we trained the multimodal models for supervised sentiment classification (Multi-Embedding \& Residual Network) and data reconstruction (Bimodal Autoencoder). The proposed approaches were trained on the train set, and we monitored the \emph{validation loss} in terms of performance tracking. Afterwards, we applied a 10-fold stratified cross-validation {\cite{krstajic2014cross}} ({Table \ref{table:fea_train}}) on the training data using the features extracted per multimodal models. 
\begin{table}[htbp]
\caption{Features Extraction Train Set Results}
\centering
    \centering
  \begin{tabular}{|c|c|c|c|c|}
    \cline{2-5}
    \multicolumn{1}{c|}{} & Mean & Min & Max & Std \\ \hline
    LinearSVC-ME & 26.40   & 16.46   & 33.53 & $\pm$ 5.3 \\ \hline
    KNN1-ME & \textbf{33.24}   & 27.41   & \textbf{39.16} & $\pm$ 2.9 \\ \hline
    KNN3-ME & \emph{32.22} & \emph{29.19}   & \emph{35.26} & $\pm$ 2.1 \\ \hline
    KNN5-ME & 30.52 & 28.93  & 33.48 & $\pm$ \emph{1.3} \\ \hline
    DT-ME & 31.54   & \textbf{29.44}   & 34.00 & $\pm$ 1.5\\ \hline
    RF-ME & 25.70  & 25.05  & 26.96 & $\pm$ \textbf{0.6}\\ \hline
    GB-ME & 31.09  & 28.82  & 33.93 & $\pm$ 1.5\\ \specialrule{.2em}{.1em}{.1em} 
    LinearSVC-BiAE & \textbf{33.60}  & \textbf{30.32}  & \textbf{36.40} & $\pm$ 2.0\\ \hline
    KNN1-BiAE & \emph{33.08}   & 29.74   & 35.58 & $\pm$ 1.9 \\ \hline
    KNN3-BiAE & 31.42 & 28.08 & 33.30 & $\pm$ 2.0\\ \hline
    KNN5-BiAE & 31.10 & 27.90 & 34.11 & $\pm$ 1.9\\ \hline
    DT-BiAE & 32.23 & \emph{29.93} & \emph{36.11} & $\pm$ 2.0\\ \hline 
    RF-BiAE & 25.15 & 24.7 & 25.99 & $\pm$ \textbf{0.4}\\ \hline
    GB-BiAE & 31.70  & 29.45  & 34.63 & $\pm$ \emph{1.6}\\ \specialrule{.2em}{.1em}{.1em} 
    LinearSVC-RB1 & 25.21 & 17.07 & 32.27 & $\pm$ 4.3\\ \hline
    KNN1-RB1 & \textbf{33.44}   & \textbf{30.73}   & \textbf{35.59} & $\pm$ 1.7 \\ \hline
    KNN3-RB1 & 31.08 & 27.55 & 34.22 & $\pm$ 2.0\\ \hline
    KNN5-RB1 & 29.94 & \emph{28.25} & 32.73 & $\pm$ \emph{1.2}\\ \hline
    DT-RB1 & \emph{31.52} & 27.57 & \emph{34.70} & $\pm$ 2.3\\ \hline
    RF-RB1 & 25.23 & 24.72 & 26.23 & $\pm$ \textbf{0.5}\\ \hline
    GB-RB1 & 30.02 & 26.45 & 32.68 & $\pm$ 2.0 \\ \specialrule{.2em}{.1em}{.1em} 
    LinearSVC-RB2 & 21.34 & 15.96 & 28.95 & $\pm$ 6.8\\ \hline
    KNN1-RB2 & \textbf{32.91}   & \textbf{30.28}   & \textbf{35.17} & $\pm$ 1.6 \\ \hline
    KNN3-RB2 & 30.50 & \emph{28.99} & 31.55 & $\pm$ 1.0\\ \hline
    KNN5-RB2 & \emph{31.09}  & 28.82  & \emph{33.93} & $\pm$ \emph{0.8}\\ \hline
    DT-RB2 & 30.82 & 28.09 & 33.24 & $\pm$ 1.8\\ \hline
    RF-RB2 & 25.06 & 24.83 & 25.28 & $\pm$ \textbf{0.1}\\ \hline
    GB-RB2 & 30.76 & 27.14 & 32.69 & $\pm$ 1.9\\ \hline
  \end{tabular}
  \label{table:fea_train}
\end{table}

We used the same metric as the one adopted by SemEval-2020 Task 8, namely the macro F1-score, defined as the average F1-score per label, without taking into account the proportion of each class in the dataset. According to the experimental results presented in {Table \ref{table:fea_train}}, classifiers using our proposed multimodal feature extraction outperformed the 
baseline score specified by the challenge, i.e., 21.76\%. It can be seen that most of the models gave better results than the given baseline. We run further experiments by evaluating the multimodal feature extraction approach on the test data ({Table \ref{tab:test}}) to observe the generalizability of the models on unseen data.

\begin{table}[htbp]
\caption{Features Extraction Test Set Results}
\centering
    \begin{tabular}{|c|c|}
    \hline
    \textbf{Model Name} & \textbf{Macro F1} \\ \hline
    KNN1\-ME & \textbf{35.03} \\ \hline
    KNN3\-ME & \emph{32.78} \\ \hline 
    DT-ME & 30.72 \\ \hline
    GB-ME & 30.68 \\ \specialrule{.2em}{.1em}{.1em}
    LinearSVC-BiAU & \textbf{31.38} \\ \hline 
    KNN1-BiAU & 26.40 \\ \hline
    KNN3-BiAU & 29.01 \\ \hline 
    DT-BiAU & \emph{30.27} \\ \hline
    GB-BiAU & 29.00 \\ \specialrule{.2em}{.1em}{.1em}
    KNN1-RB1 & \textbf{34.93} \\ \hline
    KNN3-RB1 & \emph{33.28} \\ \hline
    KNN5-RB1 & 32.65 \\ \hline
    DT-RB1 & 30.02 \\ \hline
GB-RB1 & 31.85 \\ \specialrule{.2em}{.1em}{.1em}
KNN1-RB2 & \textbf{35.27} \\ \hline
KNN3-RB2 & \emph{34.86} \\ \hline
KNN5-RB2 & 31.36 \\ \hline
DT-RB2 & 29.04 \\ \hline
GB-RB2 & 31.62 \\ \hline
  
  \end{tabular}
  \label{tab:test}
  \end{table}
  
Based on the experimental results presented in {Table \ref{table:fea_train}}, we can clearly state that KNN (K=1) was the only model that had a mean f1-score above \textbf{32.91\%} in the four feature extraction use cases and had the highest F1 score of \textbf{39.16\%} when trained with features generated by the multi-embedding model. It is also the KNN (K=1) with features from the 2nd residual block that achieved the highest macro F1 score of \textbf{35.27\%} on the test set as shown in {Table \ref{tab:test}}, which is an improvement of \textbf{13.51\%} compared to the SemEval 2020 baseline performance.

  \section{Conclusion}

In this paper, we proposed and presented a multimodal feature extraction approach that uses multimodal deep learning-based approaches to jointly learn and extract features from text and images. The features extracted using the proposed approach were used to train classical supervised machine learning models to perform sentiment classification for memes. The experimental results on the SemEval dataset {\cite{chhavi2020memotion}} showed that the multimodal feature extraction approach improved the performance of the classifier by more than \textbf{13.5\%}. In the future, we plan to further investigate the multimodal feature extraction methods with more complex pre-trained models and apply such multimodal feature extraction approaches to various multimodal tasks to verify their robustness and generalizability.

\section*{{Acknowledgements}}

This research is supported by the \'UNKP-21-4 New National Excellence Program of the Ministry for Innovation and Technology from the source of the National Research, Development, and Innovation Fund.  Supported by the Telekom Innovation Laboratories (T-Labs), the Research and Development unit of Deutsche Telekom.

\bibliographystyle{plain}
\bibliography{references.bib} 

\end{document}